# ENHANCING NETWORKING CIPHER
# ALGORITHMS WITH NATURAL LANGUAGE

John E. Ortega

Courant Institute of Mathematical Sciences,
New York University, New York, New York, USA

## ABSTRACT

*This work provides a survey of several networking cipher algorithms and proposes a method for integrating natural language processing (NLP) as a protective agent for them. Two main proposals are covered for the use of NLP in networking. First, NLP is considered as the weakest link in a networking encryption model; and, second, as a hefty deterrent when combined as an extra layer over what could be considered a strong type of encryption -- the stream cipher. This paper summarizes how languages can be integrated into symmetric encryption as a way to assist in the encryption of vulnerable streams that may be found under attack due to the natural frequency distribution of letters or words in a local language stream.*

## 1. INTRODUCTION

A stream cipher can be illustrated in many ways. In its purest algorithmic form, a stream cipher is a type of symmetric encryption algorithm [1]. A symmetric algorithm achieves encryption by using the same cryptographic keys in order to encrypt or decrypt a message where a shared secret is shared by the sender and the receiver. The sharing of a secret, as most of us know from typical childhood "keep a secret" games, is not secure. And, as a result of their lack of security, stream ciphers must be considered attackable and in need of a stronger defence against attacks and greater security.

Algorithms are the key to privacy but by their nature are public and easy to read. The public availability of algorithms along with the simple frequency of a local language can prove to be devastating for a stream cipher algorithmic modeller. At times, safe stream ciphering can be considered almost as an n-complete problem due to the numerous attacks that have occurred in the past towards them.

The idea that the algorithms should all be public and the secrecy should reside exclusively in the keys is called Kerckhoffs' principle, all serious cryptographers subscribe to this idea. [2] Knowing the cryptography relies on the keys as its secrecy, an attacker will often times focus on breaking the key that is generated by a key generation algorithm. Key generation algorithms are directly used in the majority of stream ciphers and can be considered the weakest link for transferring data due to the aforementioned details where secrecy lies within a key.

One way to prevent attackers from using publicized symmetric algorithm knowledge and key decryption techniques that break stream ciphers is to provide an extra layer of security on top of the currently available layers. The layer of security should be simple to understand while at the same time robust enough to be applied to any cipher stream available. Several methods [3] have been proposed and are used for strengthening security such as randomness, bit shifting, and the use of digits. Contrastingly, a common framework, while seemingly easy-to-decrypt and insecure, could be the use of a language for encryption. Most networking stream attacks, such as the commonly implemented replay attack [4], use the knowledge of the local language at hand to stage attacks; they normally do not consider the idea of another language being used as an extra layer of encryption.

Stream ciphers, as opposed to block ciphers, hide the pre-known fact that a message will be sent using the local language and; thus, are less prone to simple attacks. This paper explains several stream ciphers and how the addition of a foreign language as an extra layer on top of the current stream cipher capabilities can serve as an extra deterrent for attacks. First, a clear survey of traditional networking algorithms and their vulnerabilities is presented in Section 2. Then, Section 3 gives details on how natural language can be used for encryption in the networking algorithms. Finally, Section 4 describes the reliability and concludes on why it would be better to use natural language for network security.

## 2. STREAM CIPHERS

A stream cipher (or pseudo-random generator) is an algorithm that takes a short random string, and expands it into a much longer string, that still looks random to adversaries with limited resources. [5] Stream cipher algorithms are typically used as a mechanism for encryption on devices such as wireless routers where encryption is required in order to not expose the data packets that are being passed as messages. Since the data that is being passed back and forth is passed randomly and in real time, data transfer can be considered a stream of packets from one endpoint to another. A stream cipher specifies a device with internal memory that enciphers the $j$ digit $M_j$ of the message stream into the $j$ digit of $Cj$ of the cipher text stream by means of a function which depends on the secret key and the internal state of the stream cipher at time $j$. The sequence $Z_0, Z_1, Z_2, \ldots Z_n$ which controls the enciphering is called the *key stream* or *running key*. The deterministic automation which produces the key stream from the actual key $k$ and the internal state is called the *running-key generator*, or *key-stream generator*. [6]

The key-stream generator generates the running key sequence described above as the key stream. The key-stream generator combines digit by digit the key sequence, or the running key, on top of the plain text sequence in order to obtain the ciphered text that can be considered somewhat easy to attack due to the fact that the text, although in a ciphered format, is normally produced using letters and/or words from the local language where the data stream occurs.

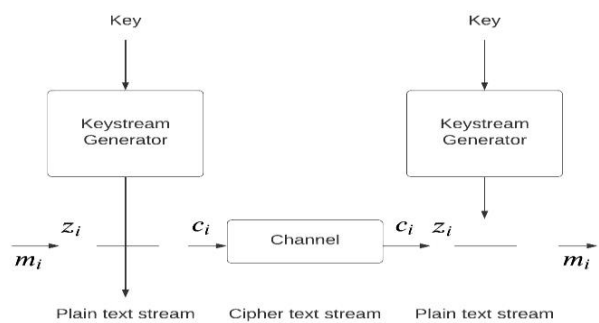

Figure 1.  Key generation in a stream cipher. [5]

The "channel" from Figure 1 above shows the typical stream flow as it passes through the stream ciphering process. The ciphered text, meant to be secure, has been found to be vulnerable to attacks due to the frequency of letters in a common alphabet such as English. [7] If the ciphered text's secret key is used more than once, while appearing random to its adversaries, it can be easily decrypted by a skilled cryptographer even though stream ciphers operate with a time-varying transformation on the individual plain text digits.

Stream ciphers depend on a pre-agreed secret for their key encryption. That idea in itself could be considered a security breach since both parties have to maintain the same secret. In this paper, the focus is on the encrypted stream and how to avoid attacks that use cryptic algorithms to decrypt the streams with prior knowledge of a particular language, especially when the same key is used more than once.

There are two major components of a stream cipher algorithm: 1) a short input string (referred to as the **key** in Figure 1 and 2) a long output string called the *key stream*. Stream ciphers can be used for shared-key encryption, by using the output stream as a one-time-pad. [1] The stream cipher can deploy random digits or letters for its encryption and decryption process, this is known as a synchronous stream ciphering process [8]. Additionally, there is another model called *self-synchronous stream ciphering* [9] that calculates ciphered digits using the previous cipher text's digits which automatically synchronize the key generator when receiving the digits.

Both stream ciphering approaches can be considered part of the stream cipher paradigm. In this paper, an additional ciphering mechanism is described to further encrypt the cipher for heightened security that uses natural language as an extra layer to the key stream.

## 2.1. Stream cipher word frequency

The random digits (numerical or alphabetical) that are formed as part of the encrypted stream in a stream cipher are usually in a local language known to the cryptographer. For that reason, a cryptographer's attempt to decrypt a stream cipher that has been created using an alphabet known by both parties and used multiple times can be considered vulnerable. An attack could be formed that uses the easy-to-discern frequency of digits that focuses on the higher occurring digits, or letters, in the local language [10].

A good example of the weakness of a stream cipher that would typically use a local alphabet for its digits is the RS4 encryption algorithm described in Section 2.2.1. Here, one can assume that the algorithm is easier to attack due to the knowledge of the language at hand. A more concrete example of local language vulnerabilities could be found in a city such as Frankfurt, Germany where an attacker would probably attack a wireless network using the German language due to the fact that more than ninety percent of Frankfurt's inhabitants use German (or Dutch) as their language of choice. On the contrary, the same principle may not be applicable for cities of higher immigration such as Miami, Florida, USA where the spoken language (Spanish) is not the official language of the country (English).

Figure 2 displays the typical frequency distribution of letters in a word of the English language and gives conclusive notions that, by using the knowledge of the digit, or letter, distribution in a language, an attacker may be able to establish an attack model paradigm with ease.

It is clear that a stream cipher whose ciphered output is generated using the English language would, judging from Figure 2 above, probably contain the letter "e" within its context. Therefore, by using the fact that certain letters are more likely to be included in a stream, attacks are normally crafted using higher occurring letters from the local language's alphabet.

A cipher that is streamed, specifically a streaming cipher that uses the same key and input data will produce identical key streams if used with the same key and input data over successive operations. Since the key stream is frequently combined with plain text using an invertible operation, this means that successive cipher texts can be combined to produce a combination of the plain text. [1] That makes an attack, such as a replay attack [4], a good candidate for attack because one could identify the commonality of a repeating stream using easy-to-obtain tools such

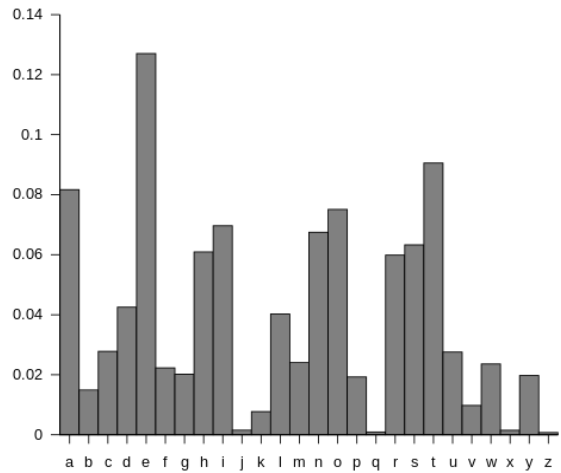

Figure 2. English language letter frequency. [11]

As Wireshark [12] and Aircrack-Ng [13]. A cryptographer could use those tools and a Wi-Fi stream to apply a special type of algorithm implementing the replay functionality that combines the knowledge of local language digit occurrences and possible input phrases to stage an attack.

In addition to simple repetition detection, local language frequency gives way to a high amount of redundancy. In the case of messages with a high amount of redundancy (like in natural language or other data formats), error propagation may be sufficient to detect modifications to a streamed message, but in general an additional cryptographic operation is needed to guarantee the integrity of a message. [14] Stream ciphers are normally processed in real time and the size and quantity of data this is passed via the two endpoints of a stream are normally unknown. Ideally, the algorithm that produces ciphered text in a stream would be random enough such that simple word frequency tactics and reasoning would not be enough for a cryptographer to decrypt. However, due to the easily attainable algorithms that are highly publicized and other general factors that apply to most stream cipher algorithms, stream ciphers are still vulnerable to attack and require an extra layer of security.

## 2.2. Stream cipher vulnerability

Stream ciphers and their counterparts, block ciphers, are vulnerable due to word frequency probability, local language use, and repetitiveness. The stream cipher key is dependent on the key generator which may produce output of a particular stream cipher that could be considered less weak due to its key. If a key has been generated using a weak algorithm, attacks can be executed with ease. Since many of the key generation algorithms are already published, certain algorithms have been proven to be more vulnerable.

### 2.2.1. RC4 algorithms in stream ciphers

In a strong key stream generator, each bit of the output will depend on the entire key for its value, and the relationship between the key and a given bit (or set of bits) should be extremely complicated. According to [15], the most widely used stream cipher is the RC4 stream cipher. RC4 is currently found in various applications. In stream cipher context, RC4 can be commonly found in a wireless protocol called wired equivalent privacy (WEP). WEP has already been considered a vulnerable protocol due to its stream cipher key vulnerability; newer protocols such as Wi-Fi protected access (WPA) have already been introduced to replace WEP. WEP is especially vulnerable when the beginning of the output key stream is not discarded, or non-random or related keys are used; some ways of using RC4 can lead to very insecure cryptosystems. [16]

RC4 generates a key stream using an internal state algorithm that has a permutation of 256 possible bytes with two 8-bit pointers. The pointers randomly swap bytes pointed to in order to XOR message bytes. RC4 can be considered a simple and quite elegant algorithm. Nonetheless, its simplicity makes it vulnerable to attacks such as the bit-flipping attack that uses the knowledge of the algorithm to decipher streamed text. It can be understood from Figure 3 that an RC4 application can be deciphered by knowledge of the algorithm easily found on the internet or other publications. A denial of service attack (DOS) [17] could be used to insert plain text that would produce a predictable output exposing the stream cipher's algorithm and, thus, makes it easier for an attacker to attack stream ciphered text. For example, previous work [18] presented an analysis of an RC4 stream cipher showing more correlations between the RC4 key stream and the key and was able to crack an RC4 encrypted algorithm for WEP in under a minute.

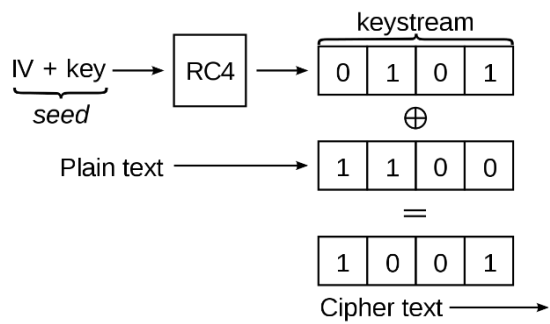

Figure 3. RC4 Stream Ciphers for WEP. [16]

RC4 attacks are now commonplace and almost any primitive hacker can use the knowledge of algorithms such as the RC4 algorithm to attack a stream such as the streams found in wireless WEP technologies for many commonly sold routers on the market. Many variations of the RC4 algorithm have been implemented and, unfortunately, successfully decrypted without the knowledge of the key. The RC4 creates a one-time key of about 24 bits for its security. 24 bit length really cannot be considered safe. The fact that the RC4 algorithm is readily available combined with its key shortness and use of local language digits make it highly vulnerable.

### 2.2.2. LFSR algorithms in stream ciphers

The LFSR (Linear Feedback Shift Register) algorithm [19] is yet another, considerably insecure, algorithm that can be used in stream ciphers to generate a key. LFSR depends on a previous state by applying a linear function to it. The most common linear function is to take the previous state's bit pattern and XOR it with some bits to modify the overall state. LFSR eventually repeats because its registers have a finite number of states and, due to the states finiteness, could be considered less secure when states are cycled repeatedly. Nonetheless, if a LFSR algorithm is chosen with a strategic security plan in mind, it could appear randomly acyclic when under attack. LFSRs have long been used as pseudo-random number generators for use in stream ciphers (especially in military cryptography), due to the ease of construction from simple electromechanical or electronic circuits, long periods, and very long periods, and very uniformly distributed output streams. [19] A skilled attacker could decrypt an LSFR quite easily using output text combined with the simulation of a receiver to gain access to encrypted information. One such attack is known as the correlation attack [20].

A correlation attack can be devised to understand the Boolean, cyclic nature of the LFSR algorithm. Predictive possibility tables can be drawn that take the possible input and output in order for the hijacker to be able to decrypt the stream cipher using Boolean logic like Figure 4. So, the decryption would intercept the stream cipher, apply the key stream generation algorithm table using statistical probability, and gain access to the stream. In order to statistically decrypt a stream cipher algorithm, the cryptographer would only have to apply the correlative technique in a key generation algorithm with an algorithm such as the Geffe generator. [21]

**Boolean function output table**

| $x_1$ | $x_2$ | $x_3$ | $F(x_1, x_2, x_3)$ |
|-------|-------|-------|---------------------|
| 0 | 0 | 0 | 0 |
| 0 | 0 | 1 | 1 |
| 0 | 1 | 0 | 0 |
| 0 | 1 | 1 | 1 |
| 1 | 0 | 0 | 0 |
| 1 | 0 | 1 | 0 |
| 1 | 1 | 0 | 1 |
| 1 | 1 | 1 | 1 |

Figure 4.  A Boolean table for a correlation attack. [21]

If the stream cipher algorithm is implemented using LFSR, the key stream may be too vulnerable and easy to attack, even with another layer of non-local language applied. While a natural language layer could be applied to a stream cipher with LFSR, the simple fact of the repetitiveness in LFSRs cycle make it easier to attain the correct keys. Nonetheless, if one could find a correlation between the output of one of the shift registers and the key stream, then one can try to find the initial state of this LFSR independently of the other LFSRs. [22] Correlation attacks are the most common way to attack LFSR key generations and serve as an example of the weakness of stream ciphers. Correlation attacks can be considered extremely dangerous and stream ciphers extremely susceptible; extreme care must be taken when designing stream ciphers in order to protect against correlation attacks.

## 3. NATURAL LANGUAGE ENCRYPTION

Natural languages are based on the day-to-day conversations that we experience and can be considered as pieces of information that help us as humans to communicate more effectively within our domain. Natural languages are governed by implied rules for which natural selection inherently defines. [23] While we can attempt to define those rules using techniques such as finite state transducers (FST) [24], it can be assumed that natural language rules are nearly impossible to approximate via mathematics or grammatical structure, at least with one-hundred percent accuracy. This motivates the study of their use in cryptography as a stronger cipher because they are complex and difficult to solve even by those highly trained in statistical digit, or letter, probability. Overall, it makes practical sense that a key generated with an extra layer of natural language may be more secure due to its grammatical and mathematical incorrectness that make prediction of the key more complex.

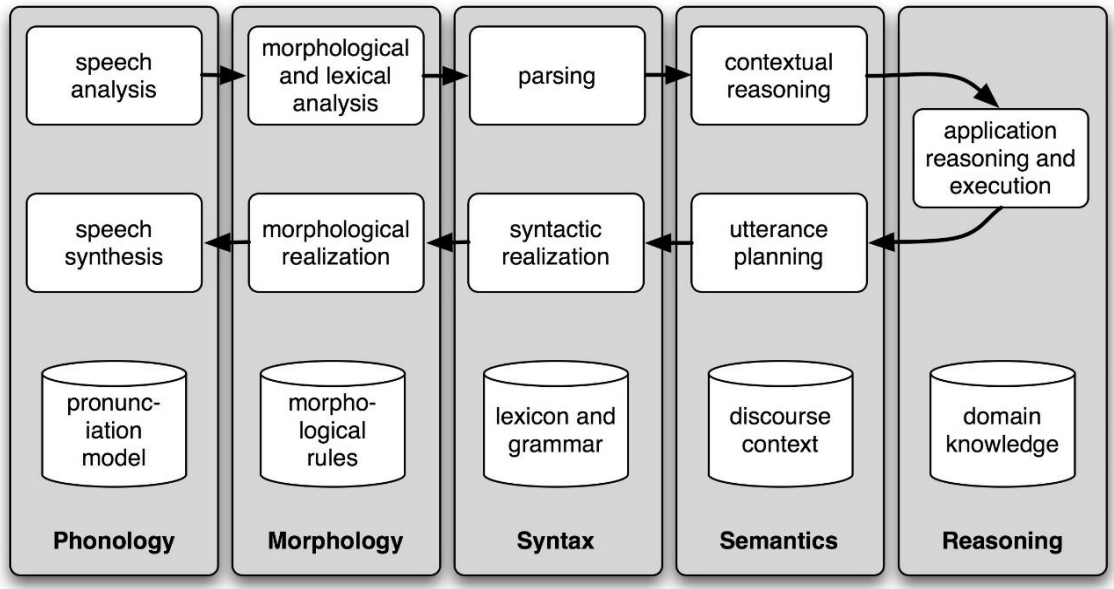

Figure 5.  NLP in software. [26]

NLP is the process of a computer extracting meaningful information from natural language input and/or producing natural language output. NLP, as the complexity of Figure 5 shows, is often considered the problem and not the solution due to the difficulty of the task of accepting natural input and producing natural output that are governed by implicit language grammatical models that may not be traceable to any group of persons. Notwithstanding, if a stream cipher is created by the implementation of a natural language that is typically spoken where the stream is being transmitted, the likelihood that the stream can be decrypted using a local natural language is higher than if it were to use a non-local language. In the following section, an introduction to the idea of encryption by using an atypical natural language is proposed.

## 3.1. Plain text language encoding

A stream cipher that is used for encoding performs its encryption at the level of individual letters or bits. Typically, a cipher, whether a stream or block cipher, uses *plain text* letters to encrypt a message. A cryptographer is considered an expert at decoding *plain text* letters. It is not a surprise, then, that plain texts are often used as targets for decryption algorithms that a cryptographer may routinely use. [10] Plain text taken from everyday sources such as newspapers, recorded telephone conversations, and wireless traffic can be considered a prime target for an attack. The knowledge that a specific target may be written in plain text combined with the fact that a target's implemented language is probably the most common language used within the target's geographic location allows cryptographers to devise plain-text algorithms using bits or letters from the local language. For example, the following scheme could be used as a way of encrypting letters in English:

A B C D E F GH I J KLMNOPQR S T UVWXYZ

QWE R T YU I O P A S D F G H J K L Z X C V B NM

This general system, according to Stallings [2], is called a **mono-alphabetic substitution**, with the key being the 26-letter string corresponding to the full alphabet. The encryption key in this example is "QWERTYUIOPASDFGHJKLZXCVBNM". For the key above, the plain text word: "ATTACK" would be transformed into the ciphered text word: "QZZQEA". Plain text can be described as the typical writings that we see written in the media that surrounds us and is often near grammatically correct. In order to understand plain text, the plain text's reader would have to have a basic knowledge of grammatical rules that govern the language that the plain text is written in. Stream cipher encoding which uses plain text is insecure when using a locally spoken language of a cryptographer. If the plain text is encoded using a highly redundant language -- such as English or any other natural language -- it can be extracted without knowledge of the key. [1] Ideally, if a sender and a receiver would like to communicate using ciphers and plain text, an encrypted layer must be applied to the plain text in order to make the cipher less vulnerable to attack. One such case where plain text was found to be undecipherable is a study on the Al-Qaeda group in the United Kingdom that used a combination of known natural languages from countries where the group exists such as Pakistan, Yemen, and Sudan. It was almost impossible for the local cryptographers who were accustomed to decrypting messages sent in the native local language, English, to decrypt messages encrypted with natural languages from other countries. [25] The encoded messages were finally decrypted by employing cryptographers from the aforementioned countries. Between them, the code-breakers spoke all the dialects that form the basis for the code. Several of them have high-value skills in computer technology. The local language, native to the Al-Qaeda group, was used as a way of encrypting plain-text messages that could not be understood by the local, mostly English native speakers, inhabitants. Plain text can seem somewhat simple to decrypt. But, if the plain text is written in a language that is not known to the reader and if that language is written in a natural (unstructured) form, it would be much more difficult to decrypt.

## 3.2. Natural language layer for ciphers

The presence of a natural language can be seen as the weak link of a stream or block cipher. While it may be difficult to determine the text of an encrypted message, given the natural language of a base encryption, a cryptographer can use word frequency algorithms, such as the Berlekamp-Massey Algorithm [27], to exploit one weakness in the decryption process. In that respect, the use of NLP can be seen as a weakness in stream ciphers; however, NLP can also be used for heightened security.

Alternative language constructs can be used as a way of obfuscating encrypted keys. In order to hide the keys, a layer of encryption for increased security can be applied in a language spoken by non-natives to enhance the quality of the algorithm. In NLP, the term "noise" can be defined as the extra phonetics or disturbance inherent to a language that makes the language hard to understand. Languages such as German could be considered "noisy" forms of the English language. [28] With sufficient distortion, or noise, a language can be undecipherable and nearly impossible to dissemble. One may consider this technique as a form of "scrambling" [29]. For example, in the United States, it is known that the Federal Bureau of Investigation has scrambled mobile phone signals when conducting investigations. [30] The noise that one hears when a mobile phone signal is scrambled makes conversations nearly impossible to understand.

This paper proposes the addition of natural language to block and stream ciphers by using a non-native "noisy" layer to scramble text in an encrypted message to the point where a local cryptographer would have a hard time decrypting the message, similar to the scrambled mobile phone message described in the previous paragraph. With additional use of a foreign language to scramble the text, an attacking cryptographer would first have to decrypt a message and then translate it, the translation would be in two or more languages make it very difficult for the most state-of-the-art machine translation systems like those from Google.

Since translation of two or more encrypted languages added as layers to stream ciphers would require that a parallel key is known by the sender and receiver. In this proposal, the parallel key is combined with a non-native language which is considered to be the "noise" of an already encrypted stream. If the noise caused by the encrypted natural language is sufficient enough to scramble encrypted messages, the type of security can be considered an addition to current standardized layers. One example of how this has been done in the past is the use of language mixing by terrorist. [31]

Consider the following example:

> **native language**: bob is a joker
> **simple encryption algorithm**: b=a, o=c, i=r, s=z, a=q, j=g, k=e, e=x, r=t
> **result**: aca rz q gcext

For a cryptographer, the example above would take seconds to decrypt. But, if an additional language was added as an extra layer of encryption, and if the language was a mixture of two or more languages, the message would be tougher to decrypt. Below is the same example using *Spanglish*, the mixture of Spanish with English, a language without official rules spoken in several parts of the world. [32]

> **mixed language**: bob es un joker
> **simple encryption algorithm**: b=a, o=c, i=r, s=z, a=q, j=g, k=e, e=x, r=t, u=h, n=l
> **result**: aca xz hl gcext

While the example above is simple in nature, the decryption technique is more difficult to decipher due to the language not only lacking grammatical sense but also having no meaning after decryption. An attack on the encrypted text above, for example, would be difficult for a person who deals only in English. Additionally, if we assume that the cryptographer was from a country where English and Spanish were not spoken (Russia for example), the decryption above would be even more difficult.

The idea of using natural language in stream ciphers will motivate cryptographers to break it, and if they break it, the stream will be hard to understand because of the ignorance of the natural language that is used in the cipher i.e. ARABIC, Chinese or Japanese, Italian or Greek languages[33]. In order to apply a natural language to a stream cipher, a dependency must be established and the encoding language set as a part of the encryption. The application of the language on top of the stream layer requires that a Unicode representation deemed as input for the second language is created. After the representation has been combined to the stream, an XOR operation is performed on the binary Unicode representation of the input in the second natural language and a binary key is then used to generate an encrypted output. Decryption is finalized by the receiving end using the reverse order.

While NLP can be used as a key deterrent against attacks, it is still not full proof. It is important, for a better level of security, that the generated keys not be repeated twice. Repetition avoidance applies to stream ciphers specifically because of the encryption cycle that occurs. It would also be wise that the stream cipher's encryption algorithm and its language counterpart use languages that are not so typical to a specific region. For example, if a key generator algorithm created for a wireless router is made in Spain, it would not be wise to create the ciphered text using an algorithm that translates to a nearly similar alphabet such as Spain's neighbouring France.

It can be noted that, by frequency alone, stream ciphers are considered vulnerable. In some ciphers, such properties of the natural language plain text are preserved in the cipher text, and these patterns have the potential to be exploited in a cipher text-only attack. Language models typically written in published algorithms can be trained to learn ciphers. While research is still ongoing, some language algorithms can learn by repetition. Therefore, the pure repetitiveness of certain words such as the article "the"' in English can serve as a weak point in a stream cipher text encryption. Cryptographers dedicate themselves to finding patterns in common texts that render symbolic patterns. By applying the NLP technique described here, decryption becomes more difficult due to the language barrier that a cryptographer would probably display. Contrastingly, multi-lingual cryptographers are more likely to find patterns in ciphered texts that have been encrypted with non-native languages due to the fact that they are probably more likely to have seen specific data points within language patterns that serve as key indicators that a stream may have been encrypted using another language.

The insertion of a distinct language in a stream is not difficult to perform. The most important role that language plays in the stream cipher is the protective role of defence. As is typical in stream ciphers, both the sender and receiver must be aware of the language applied and its rules should be made clear before a key is generated. When applying the XOR described above as a binary set, if one of the words does not match a set pattern, the decryption algorithm may be thrown off and more difficult to read. While this may sound simple to do, local languages, by their sheer use, are less likely to be bound by rules which make them less useful in general. Regardless, if a common language can be understood in a local area, rules can be applied to inject the proper encryption. The parallel key (along with key stream bits) for this type of ciphers can be the languages name itself or other world of common interest between two parties may be used[33].

The XOR operation can be considered the single most important part of applying a NLP technique to a stream. An XOR operation is also a key focus of attackers. When adding the language in as an extra layer of protection, the key generation algorithm must be careful that a replay attack can't reproduce through redundancy techniques a way of combining series of messages. For that reason, it is more secure to add a non-local language into the XOR operation. Randomness plays an important key in any key generation technique for stream ciphers. Hence, a naturally spoken language should be clearly known by both the sending and receiving algorithms in order to avoid simplistic yet meaningful collisions that can be translated using a key deciphering algorithm.

The principle vulnerability in a stream cipher, and the reason why the XOR operation is the most important, is the frequency at which letters or symbols occur within the encryption language. The final binary added on as a layer discussed in this paper should help to disqualify stream ciphering encryption detection algorithms. The likelihood of attack would highly decrease if a key is created with high security by using a key that is not repeated and random along with the extra layer of security that languages provide. An attacker would have to have great knowledge of languages and decryption in order to recognize patterns that may occur; especially, if the XOR operation implies a mixture of languages similar to those used by the military originally created by native American tribes [34].

# 4. STREAM RELIABILITY AND CONCLUSION

Application and protocol designers, even those with experience and training in cryptography, cannot be expected to always identify accurately the requirements that must be met for a mode to be used securely or the conditions that apply to the application at hand. As in [34], private enterprises such as Google and Microsoft receive millions of attacks a year. Whether an enterprise level user or a simple home user, network security, no matter at what level, can be attributed to a price with information containing a value. The protection of that information really depends on its value. Credit card numbers may be considered more important than a user id for an adventure gaming website. Higher valued items and messages are retrieved via network streams of data and are captured and decrypted by skilled cryptographers. The heightened sense of security towards streams must be considered important.

Attacks are direct and easy to accomplish with the current attacker tools available. Wireless WEP attacks have proven to be as simple as inserting a disc or usb into a laptop and pressing enters. Although the latest wireless networks seem to be more secure and robust, keys are retrieved through cryptology and it is inevitable that algorithms will be created to decrypt the most difficult encryption. But, if tactics such as the NLP layer described in this paper are employed, a cryptographer's job can be made considerably more difficult.

## AUTHORS

**John E. Ortega** is a PhD holder from the Universitat d'Alacant and a member of the European Association for Machine Translation. He has over 15 years of software engineering and development experience in the private sector. His main field of research is fuzzy-match repair, although he has also worked on low-resource machine translation and topic modelling. He has earned several patents for various bayesian techniques and is a professor and guest lecturer at New York and Columbia Universities.

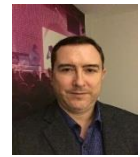